\documentclass[letterpaper, 10 pt, conference]{ieeeconf}  

\IEEEoverridecommandlockouts                              
\usepackage{graphicx} 
\usepackage{epsfig} 
\usepackage{times} 
\usepackage{amsmath} 
\usepackage{amssymb}  

\usepackage{dsfont}
\usepackage{algorithm}
\usepackage[noend]{algpseudocode}
\usepackage{hyperref}
\usepackage{csquotes}

\usepackage[dvipsnames]{xcolor}

\title{\LARGE \bf
Shaping Rewards for Reinforcement Learning with Imperfect Demonstrations using Generative Models 
}

\author{Yuchen Wu$^{1}$, Melissa Mozifian$^{2}$, and Florian Shkurti$^{1}$
\thanks{$^{1}$ Yuchen Wu ({\tt\small cheney.wu@mail.utoronto.ca}) is affiliated with the Division of Engineering Science and the University of Toronto Robotics Institute. Florian Shkurti ({\tt\small florian@cs.toronto.edu}) is affiliated with the Department of Computer Science, the University of Toronto Robotics Institute, and Vector Institute.}%
\thanks{$^{2}$ Melissa Mozifian ({\tt\small melissa.mozifian@mcgill.ca}) is affiliated with the Montreal Institute of Learning Algorithms (MILA), and the Mobile Robotics Lab (MRL) at the School of Computer Science, McGill University, Montr\'eal, Canada. This work was supported by the Natural Sciences and Engineering Research Council (NSERC).}%
}

\begin{document}

\maketitle
\thispagestyle{empty}
\pagestyle{empty}

\begin{abstract}
The potential benefits of model-free reinforcement learning to real robotics systems are limited by its uninformed exploration that leads to slow convergence, lack of data-efficiency, and unnecessary interactions with the environment. To address these drawbacks we propose a method that combines reinforcement and imitation learning by shaping the reward function with a state-and-action-dependent potential that is trained from demonstration data, using a generative model. We show that this accelerates policy learning by specifying high-value areas of the state and action space that are worth exploring first. Unlike the majority of existing methods that assume optimal demonstrations and incorporate the demonstration data as hard constraints on policy optimization, we instead incorporate demonstration data as \textit{advice} in the form of a reward shaping potential trained as a generative model of states and actions. In particular, we examine both normalizing flows and Generative Adversarial Networks to represent these potentials. We show that, unlike many existing approaches that incorporate demonstrations as hard constraints, our approach is unbiased even in the case of suboptimal and noisy demonstrations. We present an extensive range of simulations, as well as experiments on the Franka Emika 7DOF arm, to demonstrate the practicality of our method.   
\end{abstract}

\section{INTRODUCTION}

Model-free reinforcement learning has been making significant progress in complex sensorimotor control problems, particularly when optimizing end-to-end vision-based policies \cite{nvidia_driving}. The lack of need for an explicit dynamics model has nevertheless incurred a significant cost in the form of long training times, large number of interactions with the environment, and in many cases, uninformed exploration. These drawbacks often make model-free reinforcement learning impractical and unsafe to apply to real robotic systems. 

We propose a method that combines reinforcement learning (RL) with demonstrations and imitation learning (IL) in order to address these issues and accelerate the policy optimization process. Our method consists of two phases: (a) offline training of a generative model to be used as a state-action potential function $\Phi(s,a)$ for reward shaping, and (b) online RL that uses the learned potential to shape the sparse task reward, making the learning problem easier. Although generally applicable, in our case we use normalizing flows~\cite{pmlr-v37-rezende15} and Generative Adversarial Networks~\cite{NIPS2014_5423, ho_gail} learned from a small number of demonstrations for (a) and TD3~\cite{Fujimoto2018AddressingFA} for (b),
Our method provides an alternative to existing methods that combine RL with demonstrations, because it gracefully handles the case of suboptimal and noisy demonstrations. It does this by shaping the reward function to incorporate user demonstrations in the form of \textit{advice}~\cite{Wiewiora:2003:PMA:3041838.3041938} that biases the optimization process towards areas of the state-action space that the demonstrator deems high-value, without biasing the learned policy away from the optimal solution.

\textbf{Drawbacks of existing approaches:} The majority of existing works that combine RL with demonstrations~\cite{DBLP:conf/aaai/HesterVPLSPHQSO18, Vecerk2017LeveragingDF, Zhu-RSS-18} implicitly assume optimality of demonstrations, or lack of bias in the off-policy data. If the demonstration dataset is $\mathcal{D}=\{(s_i, a_i), i=1...N\}$ these methods typically solve a variant of the following problem:
\begin{eqnarray}
\underset{\theta}{\text{max}} \; \mathcal{V}^{\pi_{\theta}}(s_0) \;\; \text{subject to} \;\; \pi_{\theta}(s_i)=a_i \;\; \forall i
\label{eqn:rl_il_1}
\end{eqnarray}
where $\mathcal{V}^{\pi_{\theta}}(s_0) = \mathbb{E}_{\pi_{\theta}} \left[ \sum_{t=0}^{\infty} \gamma^t r(s_t, a_t) \; | \; s_0\right]$ is the value function corresponding to the policy $\pi_{\theta}$ and the fixed starting state $s_0$. This problem ends up being converted to one that instead has a soft regularization term for the demonstrations:
\begin{eqnarray}
\underset{\theta}{\text{max}} \; \mathcal{V}^{\pi_{\theta}}(s_0) - \beta \sum_{(s_i, a_i) \in \mathcal{D}}(\pi_{\theta}(s_i) - a_i)^2
\label{eqn:rl_il_2}
\vspace{-0.3cm}
\end{eqnarray}
There are a number of drawbacks to this formulation: 

(a) It assumes optimal demonstrations, which is often not a realistic assumption, particularly when suboptimal actions occur at states near which optimal trajectories need to pass through. In another instance of this same issue, a number of recent works, for example~\cite{DBLP:conf/aaai/HesterVPLSPHQSO18, Vecerk2017LeveragingDF}, include the demonstrations permanently in the replay buffer for off-policy RL methods, which again assumes optimality of the demonstrations. Suboptimality could be a result of the demonstrator not optimizing the same underlying reward as the RL problem, or not being an expert. Both of these possibilities are unaccounted for by the formulations in Eqns.~\ref{eqn:rl_il_1} and ~\ref{eqn:rl_il_2} and can bias the learned policy away from the optimal policy.   

(b) A number of recent papers, for example~\cite{rajeswaran}, address (a) by reducing the effect of the demonstrations over time, by replacing $\beta$ with a decreasing sequence $\beta_t$ such that $\lim_{t\to\infty} \beta_t = 0$. While this addresses the issue of suboptimality and eventually forgets the demonstrations, it introduces another design parameter, namely the speed at which the demonstrations will be forgotten.     

(c) The formulations in Eqns.~\ref{eqn:rl_il_1} and ~\ref{eqn:rl_il_2} cannot gracefully handle multi-modal action distributions at a given state. If the dataset includes $(s_i, a_i)$ and $(s_i, a_j)$ then the policy is forced to compromise by selecting the average action, which might be neither desirable nor safe. Multi-modal policies avoid this issue, but deterministic or unimodal policies do not.   

\textbf{Our contributions:} Our work extends that of Brys, Harutunyan et al.~\cite{Brys:2015:RLD:2832581.2832716}, who incorporate demonstrations via shaping potentials that, unlike the generative models that we make use of here, are not suited for handling high-dimensional structured state spaces. Our method addresses the issues above and brings the following advantages over formulations like the one in Eqn.~\ref{eqn:rl_il_2}:
\begin{itemize}
    \item It does not make any assumptions about optimality of the demonstrations, and it does not allow the demonstrations to introduce bias to the learned policy. 
    \item It does not require a forgetting schedule for the demonstrations.
    \item It can handle multi-modal demonstrations gracefully.
\end{itemize}
\noindent We demonstrate these properties via an extensive range of simulations as well as via real robot experiments on the Franka Emika 7DOF compliant arm.



\section{RELATED WORK}
There is a large number of methods augmenting RL with demonstrations, many of them in the realm of discrete MDPs and game playing, which we cannot cover here, but we include the main ideas from continuous control tasks. 

\textbf{RL + Shaping:} Our work builds upon Wiewiora et al.~\cite{Wiewiora:2003:PMA:3041838.3041938}, who showed that a state-action potential biases the $Q$-function of the original MDP, by the exact amount of the shaping potential. They introduce the notion of \textit{advice} for an RL agent. The class of shaping potentials they considered, however, was limited to discrete action and state spaces, and was not applicable to robotics, or high-dimensional systems. Our work addresses this setting by using shaping potentials that are directly trained from demonstration data via generative models. Also related is the seminal work of Ng et al.~\cite{ng1999policy}, that introduced the notion of reward shaping and the conditions under which policy invariance holds when rewards are modified. 

\textbf{RL + Optimal Demonstrations:} Deep Q-Learning from Demonstrations~\cite{DBLP:conf/aaai/HesterVPLSPHQSO18}, DDPG from Demonstrations~\cite{Vecerk2017LeveragingDF}, as well as \cite{Zhu-RSS-18}, implicitly assume optimality of the demonstrations data, unlike our work herein. In fact,~\cite{Vecerk2017LeveragingDF} assumes that the demonstration data are in the form $(s,a,r,s')$, which constrains the type of interactions that the demonstrator can have with the robot. Having access to the reward, in addition to states and actions, is difficult in scenarios where the robot is not aware of the task that the demonstrator is executing. It is also restrictive in the sense that it does not easily allow extensions, where only the states are given but not the actions. We therefore avoid assuming imitation data in that form and opt for tuples $(s,a)$ as the demonstration dataset. The notion of \textit{advice} assumes other forms, for example high-level Linear Temporal Logic formulas that guide the (discrete) RL process, as was done in~\cite{icarteusing}. SQIL~\cite{reddy} incorporates demonstrations in the replay buffer and assigns a reward of +1 to them.

\textbf{RL + Suboptimal Demonstrations:} In~\cite{rajeswaran} a schedule is followed for forgetting the demonstrations. Optimization eventually focuses on the RL objective, but only after the policy has been initialized with behavioral cloning. Abrupt transitioning was also studied in ~\cite{Cheng-UAI-18}. A shaping approach to combine imitation and RL objective is described in~\cite{sun2018truncated}, but using state-based potentials. AC-Teach~\cite{kurenkov2019acteach} handles the case of suboptimal demonstrations using an ensemble of demonstrators and Bayesian actor-critic methods. Nair et al.~\cite{nair_exploration} provide another way of not assuming optimal demonstrations, called \emph{Q-Filtering}, whereby they only keep the terms of the behavioral cloning loss for which the demonstrated action has higher $Q$ value than the action returned by the policy. A similar idea appears in~\cite{Ning2020ReinforcementLW}, as well as in~\cite{drop} where demonstrations are dropped when outside of a confidence interval.  In~\cite{Jing2020ReinforcementLF} imperfect demonstrations are handled in terms of a soft constraint, using constrained RL. \cite{rlid} provides a unified reinforcement learning objective that also handles imperfect demonstrations through soft optimality. Our method differs from this paper in that we only assume $(s,a)$ pairs, while it also assumes access to the reward. \cite{nair2020accelerating} presents a method for doing policy search while keeping it close to the behavior policy that has been learned from an offline dataset. In~\cite{Grollman2012} the case of failed demonstrations is considered, enabling the robot to learn from both successful and unsuccessful demonstrations. TAMER~\cite{tamer} and other interactive RL approaches assume a continuous user feedback mechanism that provides the RL reward, which is a significant burden on the user.   

\textbf{Batch/Offline RL:} Most works in this area ~\cite{pmlr-v97-fujimoto19a, stabilizing_q_learning, levine2020offline} aim to constrain the effect of poorly estimating $Q$ away from the collected data. Importance sampling is a commonly used way to do this, as long as one has access to the behavioral policy or can approximate it~\cite{Precup:2001, DBLP:journals/corr/MunosSHB16, DBLP:journals/corr/JiangL15, wu2019behavior, kumar2020discor}, which may lead to lack of scalability in high dimensions. Alternatively, one could force the Q-estimation to be conservative~\cite{kumar2020conservative}, but it remains unclear whether the learned policy can be easily refined via online RL. Challenges in refining a policy that has been initialized through offline RL are described in~\cite{nair2020accelerating}. 

\textbf{Residual RL:} Methods in this family~\cite{residual_rl_1, rpl, residual_rl_controller} decompose the policy into a sum of two parts, one representing prior knowledge, namely trained from demonstration data, and one residual policy that is learned through RL. 

\section{METHODOLOGY}
\subsection{State-Action, Potential-Based Reward Shaping}
Given a Markov Decision Process (MDP) $\mathcal{M}=(\mathcal{S}, \mathcal{A},\mathcal{T}, r, \gamma)$, reward shaping, as introduced in the seminal work by Ng et al \cite{ng1999policy} refers to modifying the (often sparse) reward function in order to solve another MDP $\mathcal{\widetilde{M}}=(\mathcal{S}, \mathcal{A},\mathcal{T}, \widetilde{r}, \gamma)$ such that: 
\begin{equation}
\widetilde{r}_t = r(s_t,a_t, s_{t+1}) + \gamma \Phi(s_{t+1}) - \Phi(s_t) 
\end{equation}
\noindent The function $\Phi$ is called a \textit{shaping potential}, and it is meant to make sparse reward functions more dense by providing more reward signal for the recursive computation of the state-action value function $Q(s,a)$. Ng et al. showed that the optimal value functions between the original and modified MDPs satisfies the following equation: 
\begin{equation}
\widetilde{Q}^*(s,a) + \Phi(s) = Q^*(s,a)
\end{equation}
\noindent Every optimal policy corresponding to these state-action value functions will satisfy $\pi^*(s) = \text{argmax}_a Q^*(s,a) = \text{argmax}_a \widetilde{Q}^*(s,a)=\widetilde{\pi}^*(s)$. In other words, every optimal policy for $\mathcal{M}$  will be optimal for $\mathcal{\widetilde{M}}$ and vice versa, so the optimal behavior is not affected by the shaping function, even though the value function is.

Wiewiora et al. \cite{wiewiora2003potential} showed that the shaping potential did not need to depend only on states, it could also depend on actions. The modified reward then becomes:  
\begin{equation}
\widetilde{r}_t = r(s_t,a_t, s_{t+1}) + \gamma \Phi(s_{t+1}, a_{t+1}) - \Phi(s_t, a_t) 
\label{eqn:state_action_potential}
\end{equation}
\noindent which gives rise to the following state-action value function:
\begin{equation}
\widetilde{Q}^*(s,a) + \Phi(s,a) = Q^*(s,a)
\end{equation}
\noindent In this case of state-action shaping potentials, there are no guarantees about the preservation of the optimal policy of the original MDP to the modified MDP. In fact, the optimal policy of the original MDP is 
\begin{equation}
\pi^*(s) = \text{argmax}_a \left[ \widetilde{Q}^*(s,a) + \Phi(s,a)\right]
\label{eqn:advice_policy}
\end{equation}
\noindent while the optimal policy of the modified MDP is $\widetilde{\pi}^*(s) = \text{argmax}_a \left[ \widetilde{Q}^*(s,a)\right]$, which is in general different.

Wiewiora et al. \cite{wiewiora2003potential} demonstrated potential functions for discrete state and action spaces, that were constrained to low-dimensional discrete planning problems, which are not applicable to robotics. Our paper analyzes the case where the state and action space is high-dimensional and continuous, and the shaping potential is trained via generative models, in order to support many types of demonstration data and improve the convergence properties of imitation-guided model-free RL. As long as we are able to optimally solve Eqn.~\ref{eqn:advice_policy} and $\widetilde{Q}^*(s,a)$ is well estimated, the learned policy incorporates \textit{advice} $\Phi(s,a)$, without imposing the demonstrations as hard constraints, and without introducing bias compared to the optimal policy.          

\subsection{Potentials Based On Normalizing Flows}
One of the types of state-action shaping potentials that we consider herein is a trained function ${\Phi_{\psi, c}(s,a)=c \; \text{log} \; p_{\psi}(s,a)}$ on demonstration data $\mathcal{D}=\{(s_i, a_i), i=1...N\}$. One class of generative models that have emerged in the last few years, that is able to directly optimize this log-density objective on a given dataset are \textit{normalizing flows}. 

The main idea behind this class of models is that we can use the change-of-variables formula for probabilistic distributions to transform a normal distribution (a distribution that is easy to sample) to an arbitrary distribution (from which it is difficult to sample). Given a random variable $z_0 \in \mathbb{R}^d$, such that $z_0 \sim p_0(z_0)=\mathcal{N}(0, I_d)$, and an invertible, smooth function $f: \mathbb{R}^d \rightarrow \mathbb{R}^d$ with $z_1=f(z_0)$, the change of variables formula for distributions is:
\begin{eqnarray}
    p_1(z_1) & = & p_0(z_0)\; \bigg |\text{det}\left(J_f(z_0)\right) \bigg|^{-1} 
\end{eqnarray}
\noindent Rezende and Mohamed \cite{pmlr-v37-rezende15} chained multiple of these bijective transformations to create a normalizing flow:
\begin{eqnarray}
    z_0 \sim p_0(z_0),  
    z_K & = & f_K \circ f_{K-1} \circ ... \circ f_1(z_0)  \\
    p_K(z_K) & = & p_0(z_0) \prod_{k=1}^K \; \bigg |\text{det} \left( J_{f_k}(z_{k-1}) \right) \bigg|^{-1}  
\end{eqnarray}
\noindent where $\circ$ denotes function composition. The vast majority of the recent literature on normalizing flows concerns itself with different ways to parameterize bijective functions $f_{\psi_i}(z)$ in a way that chaining multiple of them results in an expressive enough output distribution. We follow Papamakarios et al \cite{papamakarios2017masked} and we use the same bijective transformation as Masked Autoregressive Flow (MAF):
\begin{eqnarray}
    z_k^{(1)} & = & \mu_{w_{k_1}} + \text{exp}(\alpha_{v_{k_1}}) z_{k-1}^{(1)} \nonumber \\ 
    z_k^{(i)} & = & \mu_{w_{k_i}}(z_k^{(1:i-1)}) + \text{exp}(\alpha_{v_{k_i}}(z_k^{(1:i-1)})) z_{k-1}^{(i)} \quad \quad
    \label{eqn:maf_transform}
\end{eqnarray}
\noindent Here, the superscript $i \leq d$ indexes the dimensions of the random variable $z_k \in \mathbb{R}^d$, and makes the $i^{\text{th}}$ entry of the output variable depend only on entries $1...i$ of the input variable. This preserves the triangular structure of the Jacobian matrix, so the determinant remains easy to compute. The parameters of the transform $f_{\psi_k}(z_{k-1})$ described in Eqn. ~\ref{eqn:maf_transform} are $\psi_k=(w_{k_1}, v_{k_1}, ..., w_{k_d}, v_{k_d})$. The exponential term for the scaling factor is meant to ensure the positivity of standard deviation.

Training a normalizing flow is typically done via maximum likelihood estimation, by optimizing the parameters $\psi=(\psi_1, \psi_2,...,\psi_K)$, so that the log likelihood of the points in the sample dataset is maximized. In our case, we treat $z_K=(s,a)$, since we assume access to states and not high-dimensional image data. The log-likelihood objective we want to maximize is:
\begin{equation}
    \mathcal{L}(\psi, \mathcal{D}) = -\sum_{(s_i, a_i) \in \mathcal{D}} \; \sum_{k=1}^K \; \text{log} \; \bigg|\text{det}\left(J_{f_{\psi_k}}(z_{k-1})\right)\bigg| 
\end{equation}
\noindent In order to avoid learning density functions $p_K(z_K)$ that exhibit large changes whenever $z_K=(s,a)$ changes slightly, we regularize the Jacobian of the learned density with respect to its input $z_K$. Our final training cost for the shaping potential based on normalizing flows is:
\begin{equation}
    \mathcal{L}_{\text{flow}}(\psi, \mathcal{D}) = \mathcal{L}(\psi, \mathcal{D}) + \eta || \nabla_{z_K} \text{log} \; p_K(z_K) ||^2 
\end{equation}
\noindent Once the optimal parameters $\psi^*$ are identified from the training process, we use the following shaping potential:   
\begin{eqnarray}
   \Phi_{\psi^*, c}(s,a)=c \; \text{log} \; ( p_{\psi^*}(s,a) + \epsilon)
   \label{eqn:nf_potential}
\end{eqnarray}
\noindent with $z_K=(s,a)$, $c \in \mathbb{R}^{+}$ a hyperparameter, and $\epsilon$ is a small constant to prevent numerical issues and the log probability from going to negative infinity.

\textbf{Scalability:} We note that if we had chosen to make the policy input be high-dimensional, for example image-based, our current model with $z_K=(s,a)$ would be very slow to train due to the cost of evaluating the Jacobian in Eqn. \ref{eqn:nf_potential} and the autoregressive structure of the flow transform in Eqn.~\ref{eqn:maf_transform}. That said, as we will see in the experimental results section, we have used normalizing flow shaping potentials with dimension of $s,a$ being around 30 without any issues.      

\subsection{Potentials Based On Generative Adversarial Networks}
The second type of state-action shaping potentials that we consider in this paper are functions $\Phi_{\psi, c}(s,a)=c \; D_{\psi}(s,a)$, trained on demonstration data $\mathcal{D}=\{(s_i, a_i), i=1...N\}$, where $D_{\psi}(s,a)$ is the discriminator of a Generative Adversarial Network (GAN) \cite{NIPS2014_5423}. These models also include a generative model $G_{\phi}(z)=\widetilde{x}$ that accepts a noise input $z \sim \mathcal{N}(0, I_d)$ and transforms it into a more structured random variable $\widetilde{x} \in \mathbb{R}^d$.   

Training the generator and the discriminator is not done via maximum likelihood in this case, but through a minimax optimization problem. Let $p_r(x)$ be the real distribution of the random variable $x$ and $p_{\phi}(x)$ is the distribution induced by the generator. The end goal of the training process is to optimize the parameters of the generator, so that the distance between the real distribution and the generated distribution is minimized. The discriminator parameters are optimized so that its output is high on real samples and low on (fake) generated samples. 

We follow Arjovsky et al. \cite{pmlr-v70-arjovsky17a} and Gulrajani et al \cite{Gulrajani:2017} to estimate the Wasserstein-1, or Earth Mover's distance, in order to evaluate the cost of the optimal transport plan between two probability distributions $p_r$ and $p_{\phi}$:
\begin{equation}
    W(p_r, p_{\phi})=\underset{\gamma \in \Pi(p_r, p_{\phi})}{\text{inf}} \; \mathbb{E}_{(x,y) \sim \gamma(x,y)} \left[ ||x-y||\right] \quad
\end{equation}
\noindent where $\gamma(x,y)$ indicates how much mass needs to be transported from $x$ to $y$ to transform the distribution $p_r$ to $p_{\phi}$, and $\Pi(p_r, p_{\phi})$ is the set of all joint distributions, whose marginals are $p_r$ and $p_{\phi}$. Given a fixed generator $G_{\phi}$, the intractable definition above is equivalent to the more tractable one: 
\begin{equation}
    W(p_r, p_{\phi})=\underset{D \in \mathcal{F}}{\text{sup}} \; \left[ \mathbb{E}_{x \sim p_r} \left[D(x) \right]  - \mathbb{E}_{\widetilde{x} \sim p_{\phi}} \left[D(\widetilde{x}) \right]  \right]
\end{equation}
\noindent where $\mathcal{F}=\{D: \mathbb{R}^d \rightarrow \mathbb{R} \; \text{such that} \;  ||D||_L \leq 1\}$ is the set of discriminator functions with Lipschitz constant 1. Sampling from $p_{\phi}$ is done by $z \sim \mathcal{N}(0, I_d)$ and $\widetilde{x}=G_{\phi}(z)$. To impose the Lipschitz constant of 1 on the discriminator we follow WGAN-GP in Gulrajani et al. \cite{Gulrajani:2017}, and impose a soft constraint to its gradient. The approximate Wasserstein distance can be computed this way:
\begin{eqnarray}
    L_1(\psi, \phi) & = & \mathbb{E}_{x \sim p_r} \left[D_{\psi}(x) \right]  - \mathbb{E}_{\widetilde{x} \sim p_{\phi}} \left[D_{\psi}(\widetilde{x}) \right] \\
    L_2(\psi, \phi) & = & \mathbb{E}_{\widehat{x} \sim p_{\phi}} \left[ (||\nabla_{\widehat{x}} D_{\psi}(\widehat{x})|| - 1)^2\right] \\
    \widetilde{W}(p_r, p_{\phi}) & = & \underset{\psi}{\text{max}} \; L_1(\psi, \phi) - \alpha L_2(\psi, \phi) \label{eqn:approx_wasserstein}
\end{eqnarray}
\noindent where $\widehat{x}  =  \epsilon x + (1 - \epsilon) \widetilde{x}$ with $\epsilon \sim U[0,1], x \sim p_r, \widetilde{x} \sim p_{\phi}$ is used to enforce the Lipschitz constraint on samples between the real distribution and the generated distribution, since the Lipschitz constant needs to be 1 for every possible input to the discriminator.

The approximate Wasserstein distance in Eqn.~\ref{eqn:approx_wasserstein} corresponds to a fixed generator. For the generator to improve and minimize this distance, we solve the following problem:
\begin{eqnarray}
    \psi^*, \phi^* = \text{arg} \underset{\phi}{\text{min}}\underset{\psi}{\text{max}} \; L_1(\psi, \phi) - \alpha L_2(\psi, \phi) 
\end{eqnarray}
\noindent The potential then becomes $\Phi_{\psi^*, c}(s,a)=c \; D_{\psi^*}(s,a) \label{eqn:gan_potential}$ 

\noindent \textbf{Scalability:} Training the potential on high-dimensional demonstration data is scalable as GAN training has been demonstrated to produce realistic images of faces at high resolution~\cite{karras2018progressive}. 

\subsection{Combining Reinforcement and Imitation Learning via Shaping}

\begin{algorithm}
\caption{TD3 with Demonstrations via Reward Shaping}\label{alg1}
\begin{algorithmic}[1]
\Statex \textbf{Offline pre-training}
\State $\text{Collect demonstrations} \; \mathcal{D} = \{(s_i, a_i), i=1...N\} $
\State $\text{Train shaping potential} \; \Phi_{\psi^*, c}(s,a) \; \text{from Eqn.}~\ref{eqn:nf_potential} \; \text{or} \; \text{GAN}$
 \Statex \hrulefill
 \State $\text{Given MDP} \; \mathcal{M}=(\mathcal{S}, \mathcal{A},\mathcal{T}, r, \gamma)$
 \State $\text{Consider MDP} \; \mathcal{\widetilde{M}}=(\mathcal{S}, \mathcal{A},\mathcal{T}, \widetilde{r}, \gamma) \; \text{from Eqn.}~\ref{eqn:state_action_potential} \; \text{with}$
 \Statex $\widetilde{r}_t = r(s_t,a_t, s_{t+1}) + \gamma \Phi_{\psi^*,c}(s_{t+1}, a_{t+1}) - \Phi_{\psi^*, c}(s_t,a_t) $
 \Statex \hrulefill
 \Statex \textbf{TD3 training with reward shaping}
 \State $\text{Initialize two critic networks for} \; \widetilde{M}: \widetilde{Q}_{\theta_1}, \widetilde{Q}_{\theta_2}$
 \State $\text{Initialize actor network} \; \pi_{\phi}$
 \State $\text{Initialize target networks} \; \theta_1^{'} \leftarrow \theta_1, \theta_2^{'} \leftarrow \theta_2, \phi^{'} \leftarrow \phi$
 \State $\text{Initialize replay buffer} \; \mathcal{B} \; \text{to empty}$
\While{not converged}
\For{$\text{episode}\ e = {1...E}$}
\For{$\text{step}\ t = {1...T}$}
\State $\text{Apply action} \; a = \pi_{\phi}(s) + \epsilon, \epsilon \sim \mathcal{N}(0, \sigma)$ 
\State $\text{Observe reward} \; r \; \text{and new state} \; s' \; \text{from} \; \mathcal{M}$
\State $\text{Store transition tuple} \; (s, a, r, s') \; \text{in} \; \mathcal{B}$
\EndFor
\EndFor
\For {$\text{batch}\ b = {1...B}$}
\State $\text{Sample mini-batch} \; \mathcal{B}_b \; \text{of} \; (s, a, r, s') \; \text{from} \; \mathcal{B}$
\State $\text{Sample mini-batch} \; \mathcal{D}_b \; \text{of} \; (s_d, a_d) \; \text{from} \; \mathcal{D}$
\State $a' \leftarrow \pi_{\phi'} (s') + \epsilon, \epsilon \sim \text{clip}(\mathcal{N}(0, \sigma'), -\delta, \delta)$
\State $\text{Target value}\;$
\Statex $\quad \quad \quad  y = \widetilde{r} + \gamma \text{min}\{\widetilde{Q}_{\theta_1'}(s',a'), \widetilde{Q}_{\theta_2'}(s',a')\}$
\State $\text{Update critics} \; \theta_i \leftarrow \text{argmin}_{\theta_i} \sum (y - \widetilde{Q}_{\theta_i}(s,a))^2$
\If {$b \; \text{mod} \; d$}
\State $\text{Update policy}$
\State $\phi \leftarrow \text{argmax}_{\phi} \sum_{s \in \mathcal{B}_b \cup \mathcal{D}_b}[ \widetilde{Q}_{\theta_1}(s,\pi_{\phi}(s)) +$
\Statex $\quad \quad \quad \quad \quad \quad \quad \Phi_{\psi^*,c}(s,\pi_{\phi}(s)) ]$
\EndIf
\EndFor
\State $\text{Update target networks}$
\State $\theta'_i \leftarrow \tau \theta_i + (1-\tau) \theta'_i$
\State $\phi' \leftarrow \tau \phi + (1-\tau) \phi'$
\EndWhile
\end{algorithmic}
\end{algorithm}

We now show how to integrate the learned shaping potentials in a model-free reinforcement learning method. We use Twin Delayed Deterministic Policy Gradient (TD3) \cite{Fujimoto2018AddressingFA} since it is one of the best performing model-free RL methods at the time of writing. TD3 is an actor-critic method that maintains two critic networks for the $Q$ function and one actor network for the deterministic policy. The use of the double-$Q$ networks helps by reducing overestimation bias in the $Q$-function, which leads to suboptimality in the learned policy. 

\section{EVALUATION}
We evaluate our method both in simulation and on a real robot. Our aim is to clarify the following questions:
\begin{itemize}
    \item Does our method exceed the performance of (a) behavioral cloning and (b) pure RL? 
    \item Is our method robust to random seeds?
    \item Is our method robust to suboptimal demonstrations? In particular, does it do better than RL with behavioral cloning, as formulated in Eqn. ~\ref{eqn:rl_il_2}?
    \item Is our method practical on a real robot? 
\end{itemize}
\noindent We answer all these questions in the affirmative and we analyse our experimental results below. 




\subsection{Robustness to Random Seeds}

\begin{figure}[t!]
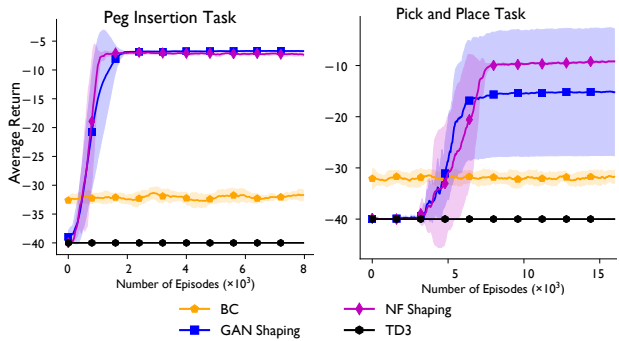

    \centering
    \includegraphics[width=0.23\textwidth]{figures/RobustToSeedsRL_fig1.pdf}
    \includegraphics[width=0.22\textwidth]{figures/RobustToSeedsRL_fig2.pdf}
    \includegraphics[width=0.22\textwidth]{figures/RobustToSeedsRL_legend.pdf}

    \caption{GAN and Normalizing Flow (NF) shaping and baseline results for \textit{peg insertion} and \textit{pick and place} tasks on the Fetch environment adopted from OpenAI Gym. The initial position of the gripper is selected randomly but at a certain distance away from the hole, and demonstrations are near-optimal. In both cases, both RL $+$ shaping methods converge to the optimal policy. TD3 fail to converge due to insufficient exploration, i.e. it never finds the goal state. Behavioral Cloning (BC) only succeeds when the arm is initialized to certain states. The empirical mean has been computed from 5 random seeds, and the error bars represent 1$\sigma$ standard deviation.}
    \label{fig:robusttoseedRL}
    \vspace{-0.3cm}
\end{figure}
The issue of robustness of policies learned via reinforcement is intricately linked to the choice of random seeds, which determine the sequence of pseudorandom number generation that will drive the exploration process, as well as the random dynamics of the environment. Henderson et al \cite{henderson2018deep} showed that many recent deep RL methods are extremely sensitive to the selection of random seeds. 

We evaluated our method on complex manipulation tasks of \textit{pick and place} and \textit{peg insertion} in simulation.

(1) \textit{peg insertion}: the end effector of the robot arm is initialized at a random location that is at a certain distance away from the hole, holding the peg. The location and orientation of the hole is fixed. A reward of $0$ is given when more than half of the peg is inserted in the hole and $-1$ otherwise.

(2) \textit{pick and place}: the object is placed at a random location that is at a certain distance away from both the end effector of the arm and the goal. The initial pose of the robot arm and the goal location are fixed. A reward of $0$ is given when the object is within a small threshold around the goal location and $-1$ otherwise.

For both environments, the episode length is set to $40$, and the environment does not terminate early.

Fig.~\ref{fig:robusttoseedRL} shows our method and  baseline results for \textit{peg insertion} and \textit{pick and place} tasks. We consider two baselines for all experiments, namely Behavioral Cloning (BC), which is pure supervised learning, and pure model-free RL (TD3) without demonstrations. All empirical results are presented with empirical mean and a single standard deviation across $5$ random seeds. The demonstration data for Fig.~\ref{fig:robusttoseedRL} are near optimal with no additional noise having been added.

Fig.~\ref{fig:robusttoseedRL} shows that while the two RL with shaping methods converge to goal, the Behavioral Cloning and pure RL method fail to explore sufficiently to find the goal area. The GAN shaping method converged to $4/5$ seeds, so the lower mean and the higher variance is due to that. 


\subsection{Robustness to Suboptimal Demonstrations}

\begin{figure}[h!]
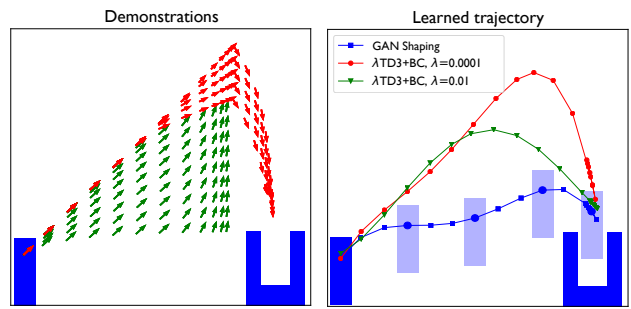

    \centering
    \includegraphics[width=0.23\textwidth]{figures/RobustToNoiseTrajectory_fig1.pdf}
    \includegraphics[width=0.23\textwidth]{figures/RobustToNoiseTrajectory_fig2.pdf}

    \vspace{-0.2cm}
    \caption{Illustration of our method's robustness to noisy demonstration data. The left figure shows the provided demonstration data to all $3$ methods on the right figure: TD3 with GAN shaping and $\lambda$TD3 + BC for $\lambda=0.0001$ and $\lambda=0.01$, which refers to the relative weight of the RL objective compared to the Behavioral Cloning (BC) objective. In this dataset, suboptimality in the demonstration data (red action vectors) is introduced by exaggerating the lift of the peg. Green action vectors on the left figure force this. Crucially, the suboptimal demonstrations are in an area of the state space where the optimal trajectory needs to pass through, so the two objectives will clash. The green curve shows the performance of the policy trained with one choice of $\lambda$ which in turns puts more emphasis on the demonstration data, which leads to convergence to a suboptimal policy. With careful tuning of $\lambda$, TD3 $+$ BC achieves suboptimal performance, whereas with GAN Shaping, the policy performs optimally. }
    \label{fig:robusttonoisetrajectory}
\end{figure}

To illustrate the sensitivity of TD3$+$BC to noise, we simplified the \textit{peg insertion} task by fixing the initial pose of the robot arm and limiting the state space to a 2D plane as shown in Figure~\ref{fig:robusttonoisetrajectory}. We provided suboptimal demonstration data that encourages the agent to lift the peg to a high location and then perform the insertion, shown as red arrows. In addition, we also included demonstration data that pushes the learned policy away from the optimal trajectory, shown as green arrows. More crucially, these suboptimal actions are given in areas of the state space where the optimal trajectory passes through, so the imitation objective is directly clashing with the RL objective.
\vspace{-0.2cm}

\begin{figure}[t!]
    \centering
    \includegraphics[width=0.23\textwidth]{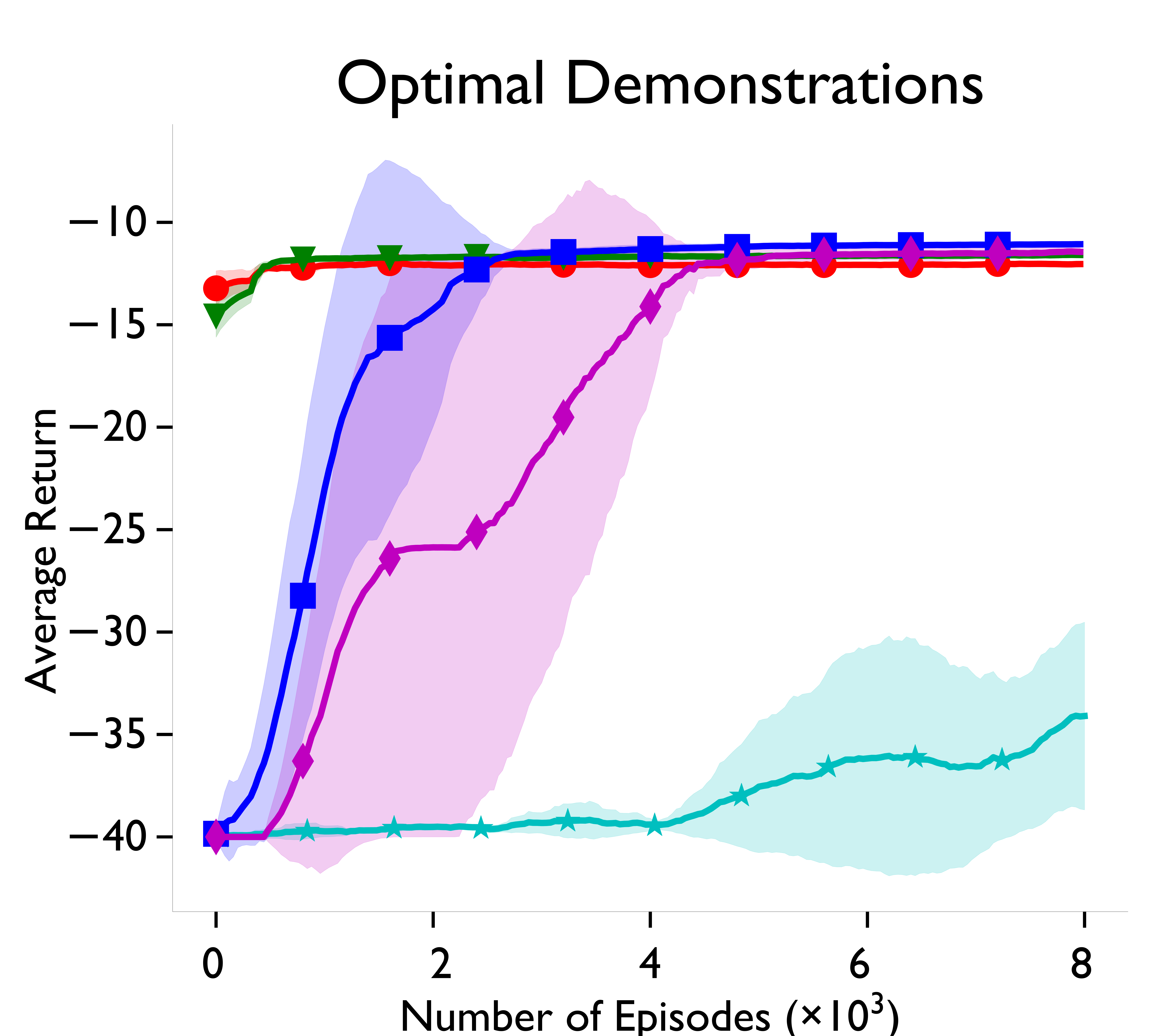}
    \includegraphics[width=0.223\textwidth]{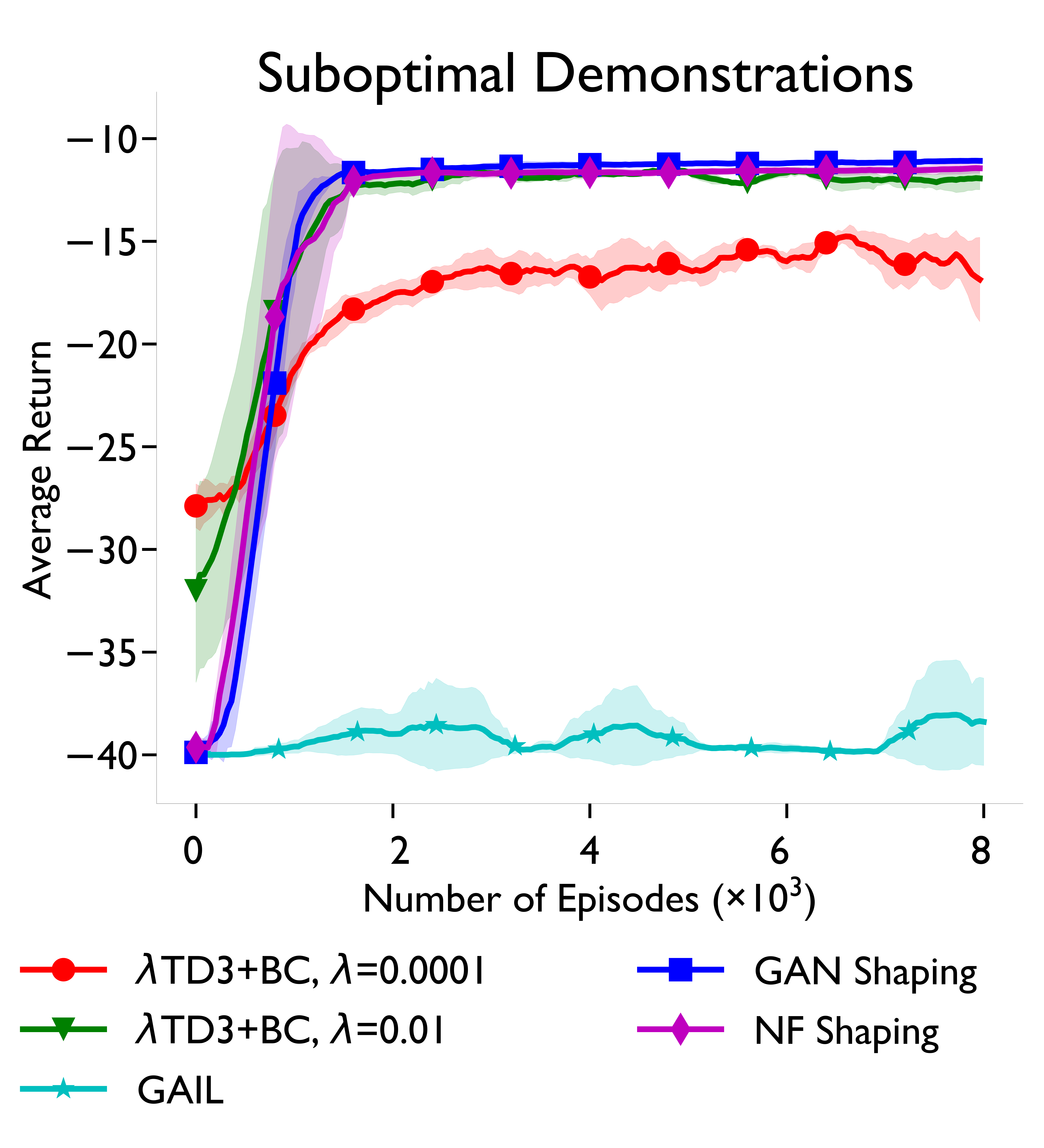}
    \includegraphics[width=0.23\textwidth]{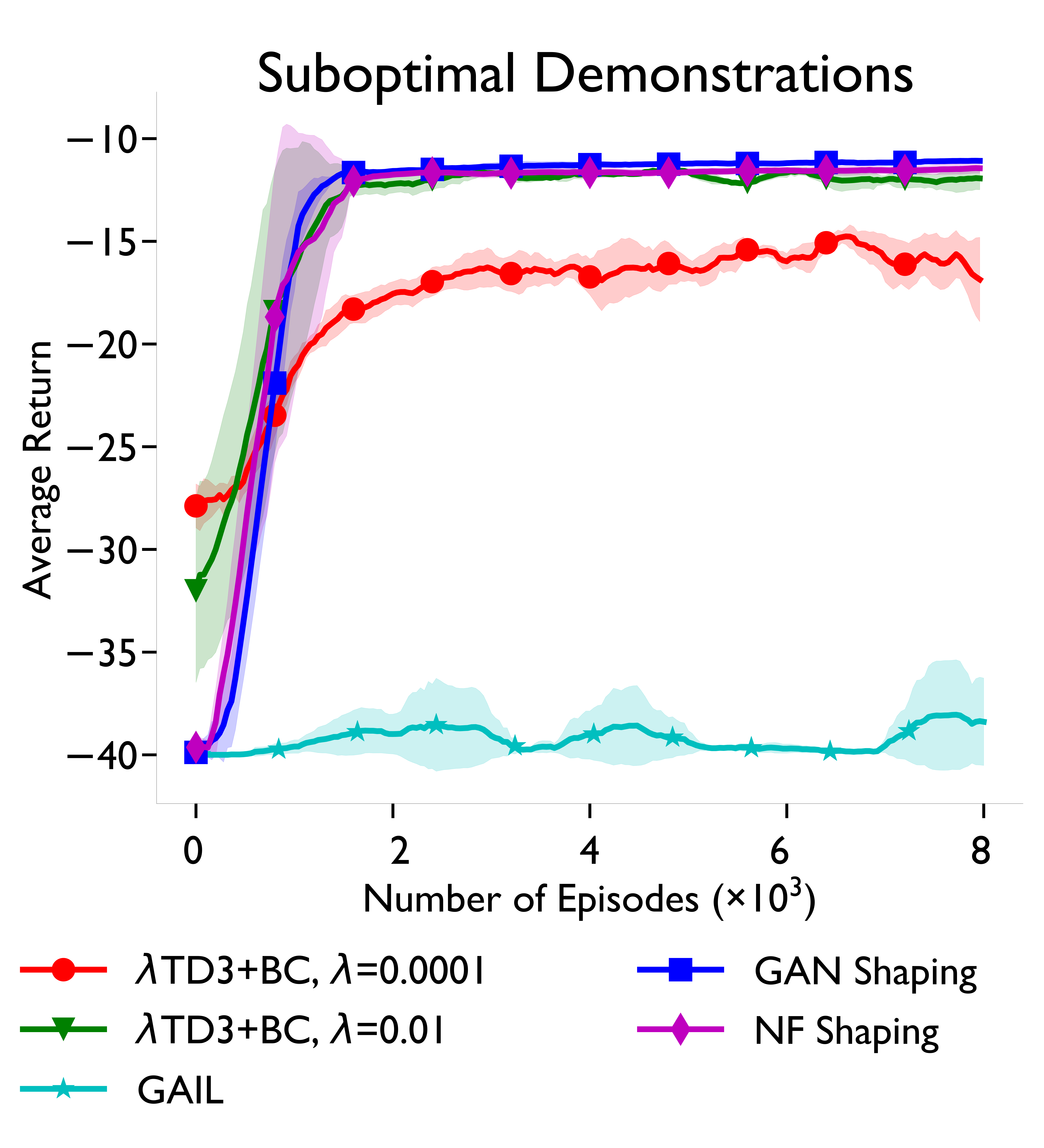}
    \caption{2D Peg insertion task. Comparison of our method that involves TD3 with shaping vs $\lambda$TD3+BC with various $\lambda$ weights, in the case of \textit{optimal demonstrations} (top) and \textit{suboptimal demonstrations} (bottom). The latter are shown in Fig.~\ref{fig:robusttonoisetrajectory}. The hyperparameter $\lambda$ refers to the relative weighting of the RL objective vs the behavioral cloning objective. These results show that $\lambda$TD3+BC is sensitive to this hyperparameter. For example, $\lambda=0.0001$ does not find the optimal policy, whereas the other methods do. Both shaping methods outperform  $\lambda$TD3+BC. GAIL, an imitation learning method, is not able to solve the task, neither with optimal nor with suboptimal demonstrations.}
    \label{fig:robusttonoiseavgreturn}
    \vspace{-0.3cm}
\end{figure}

The effect of these suboptimal demonstrations is shown in Fig.~\ref{fig:robusttonoiseavgreturn}. In particular, we see that the RL + shaping methods converge to the optimal policy, regardless of whether the demonstration data is near-optimal or not. On the other hand, the RL + Behavioral Cloning (with constraints, such as in Eqn.~\ref{eqn:rl_il_2}), is sensitive to the relative weight of the RL vs the imitation objective. When the role of the RL objective gets reduced by $\lambda=0.0001$ compared to the imitation objective, the learned policy does not manage to find the optimal solution, while for other such settings (e.g. $\lambda=0.01$) it does. This sensitivity to the relative weighting of the RL and imitation objective is undesirable, as it will affect any demonstrator forgetting schedule that adjusts this weight over time. GAIL~\cite{ho_gail}, one of the leading imitation learning methods, is unable to solve the task in the presence of imperfect demonstrations.   

\textbf{Limitations}: We note that, in general, including the potential function in the policy optimization requires tuning the regularization parameters of the generative model that change how sharply peaked the density model or the discriminator is around observed data points. In practice, this could make the optimization landscape such that it is easy to get stuck in local minima.   

\subsection{Real Robot Experiments}
For real robot experiments, we use a Franka Emika Panda 7DOF arm to perform the \textit{peg insertion} task similar to the experiments in simulation as discussed in section $A$. The learned policy controls the end effector of the robot in Cartesian velocity control mode. In order to encourage faster convergence, we fixed the initial pose of the arm and modified the reward structure such that a reward of $0$ is given when the peg is inside the hole, $-0.5$ is given when the agent is holding the peg above the hole, and $-1$ otherwise. During training, the number of episode steps is set to $100$ and episodes do not terminate early.

\setlength{\tabcolsep}{0.1em}
\def\arraystretch{0.5}
\begin{figure}
\begin{center}
\begin{tabular}[ht!]{ccc}

\includegraphics[width=0.16\textwidth]{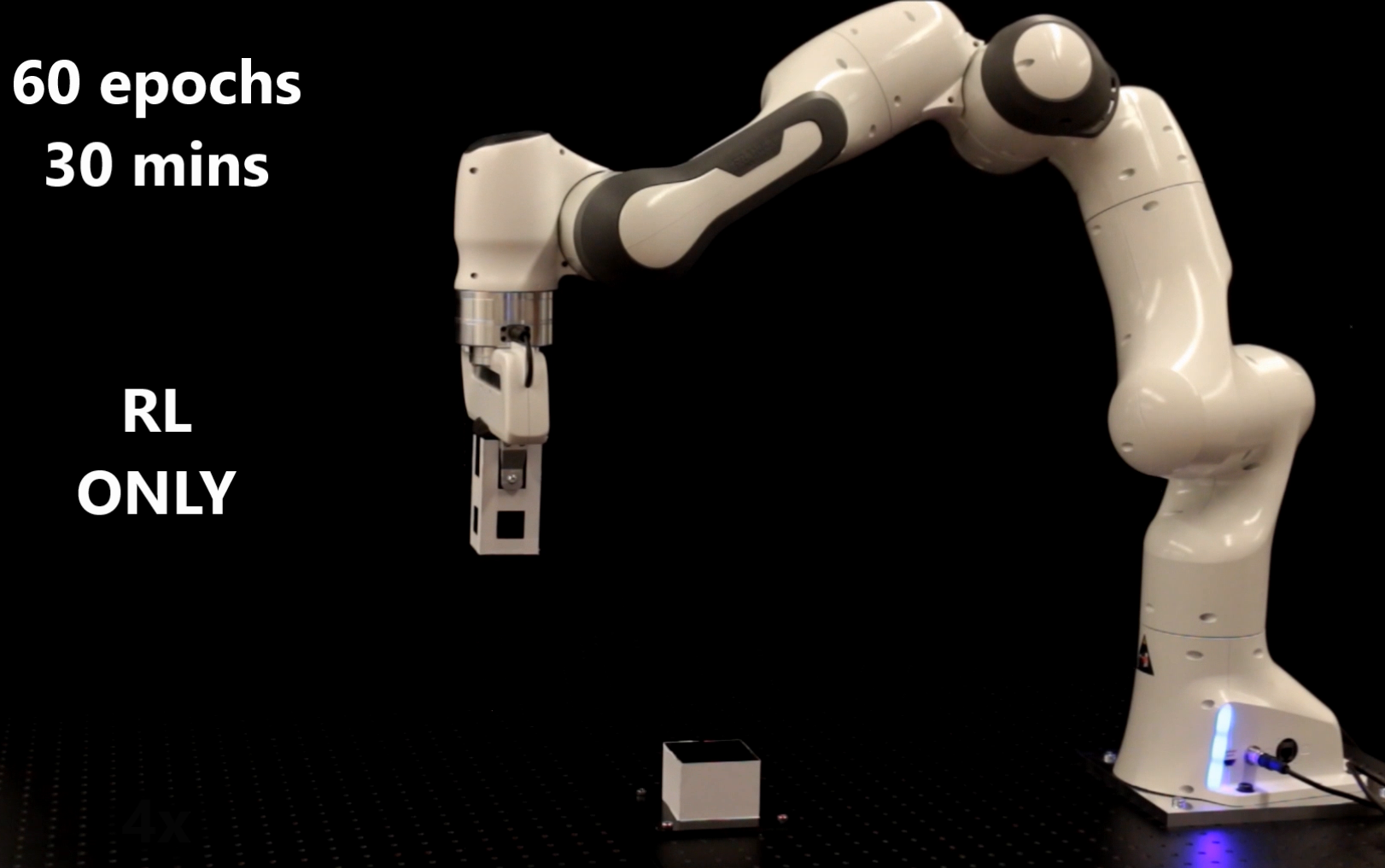} &
\includegraphics[width=0.16\textwidth]{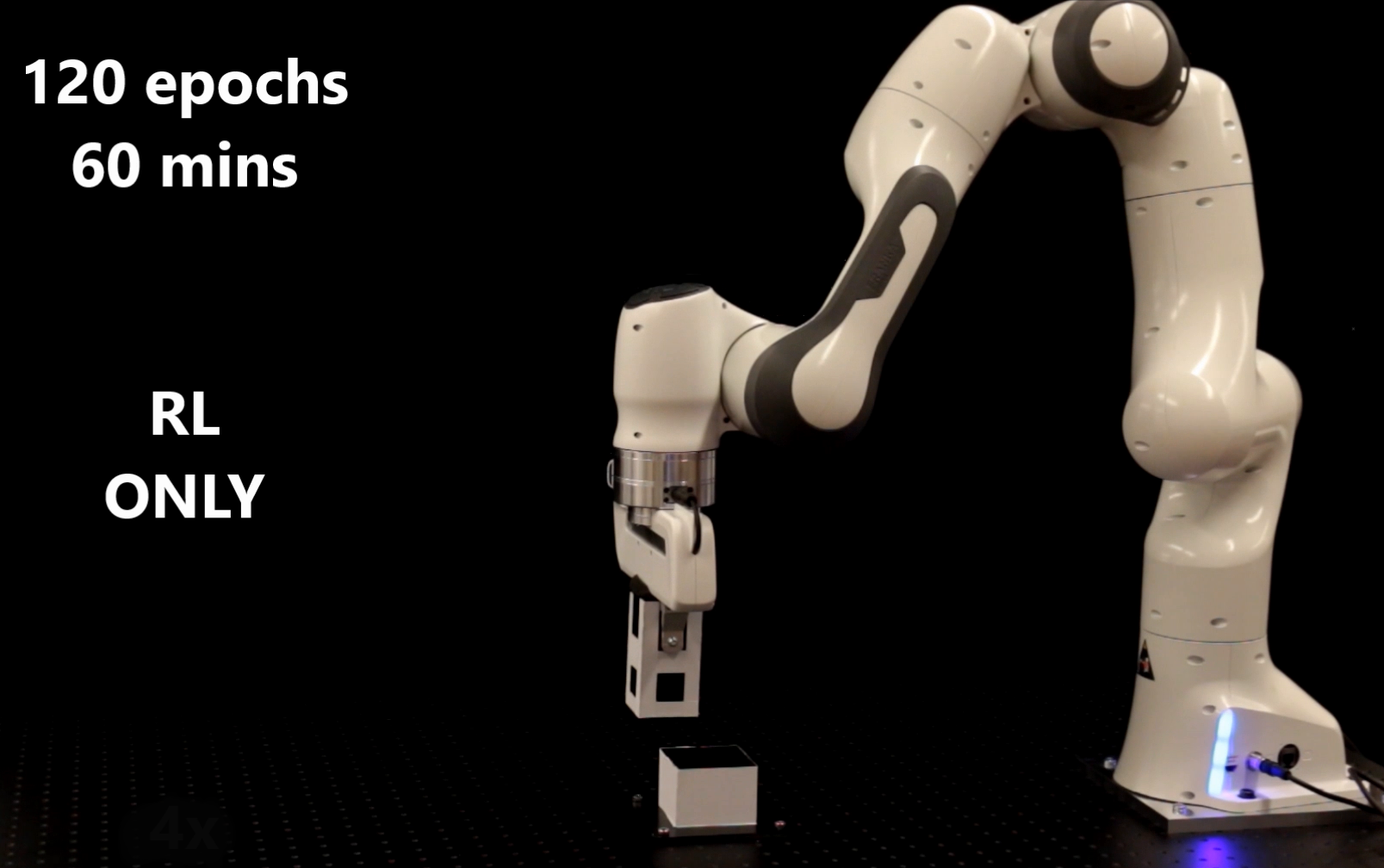} & 
\includegraphics[width=0.16\textwidth]{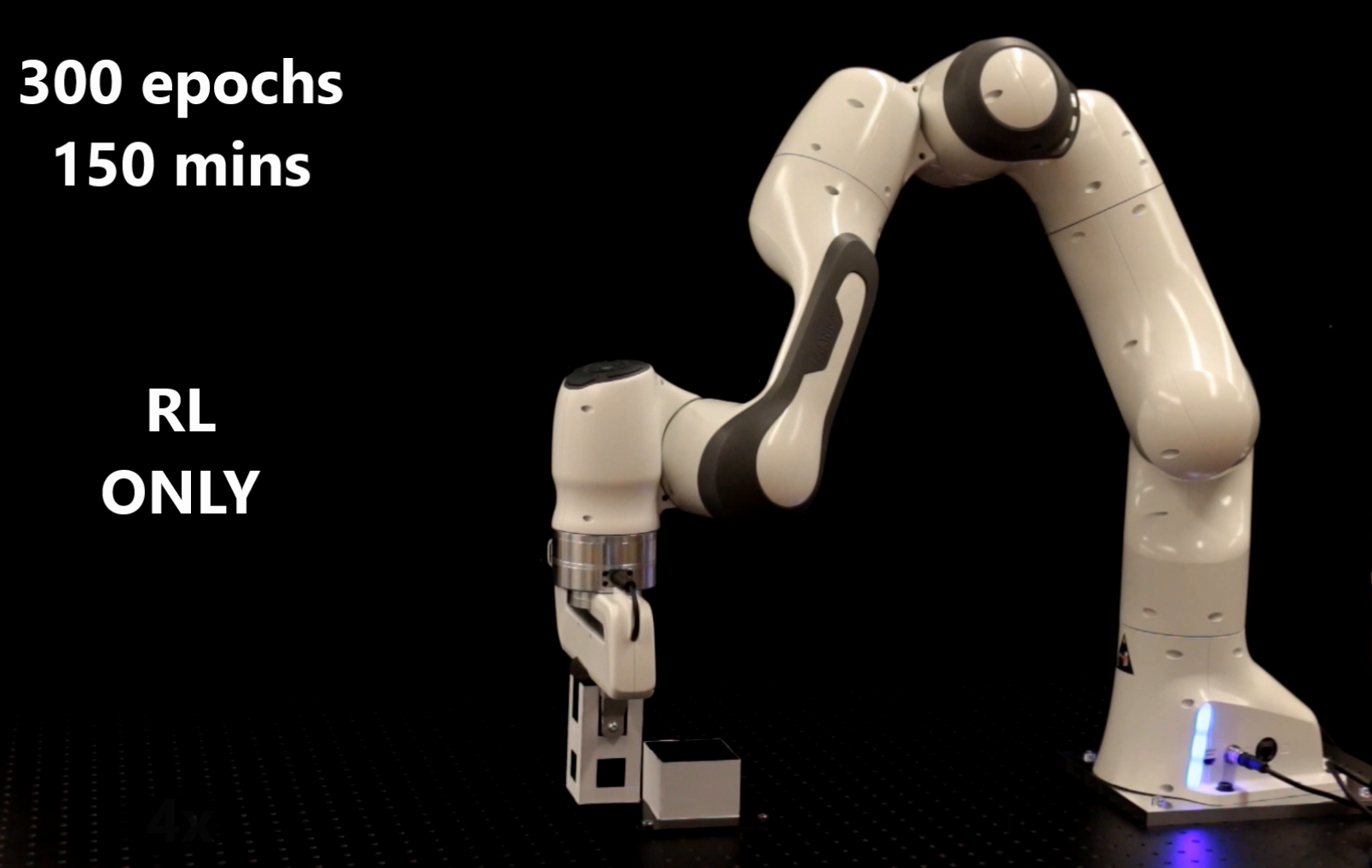} \\ 
\includegraphics[width=0.16\textwidth]{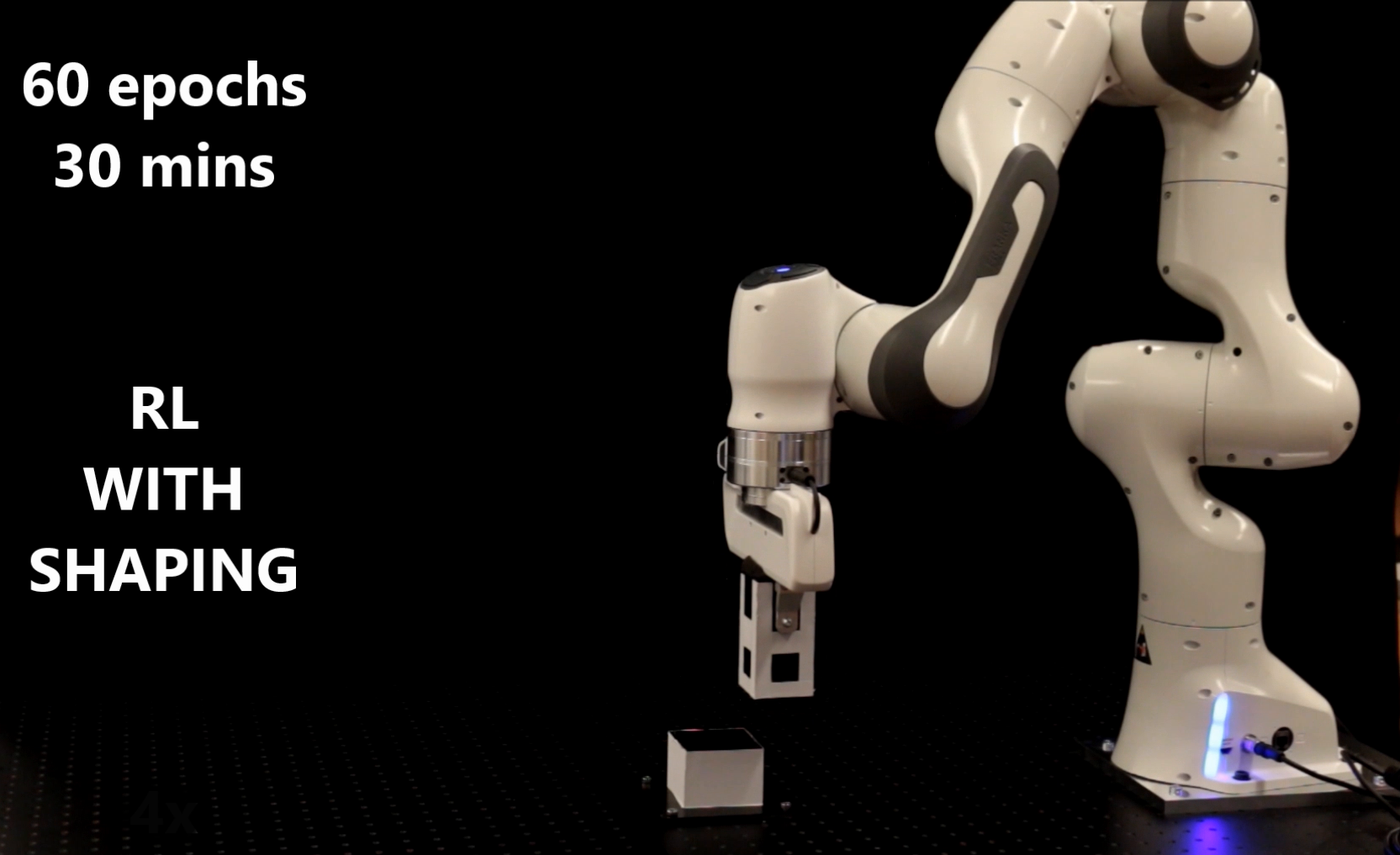} & 
\includegraphics[width=0.16\textwidth]{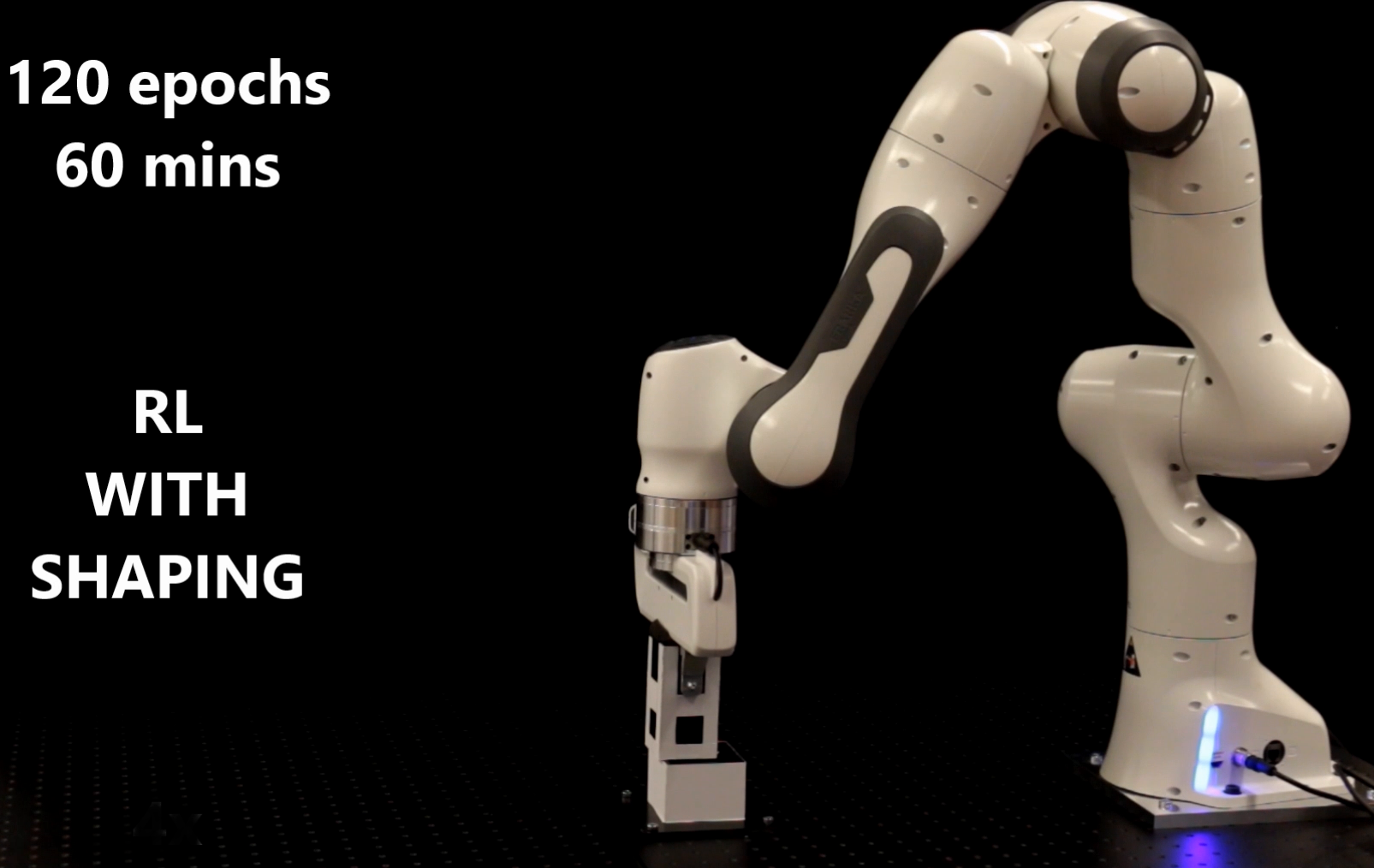} &
\includegraphics[width=0.16\textwidth]{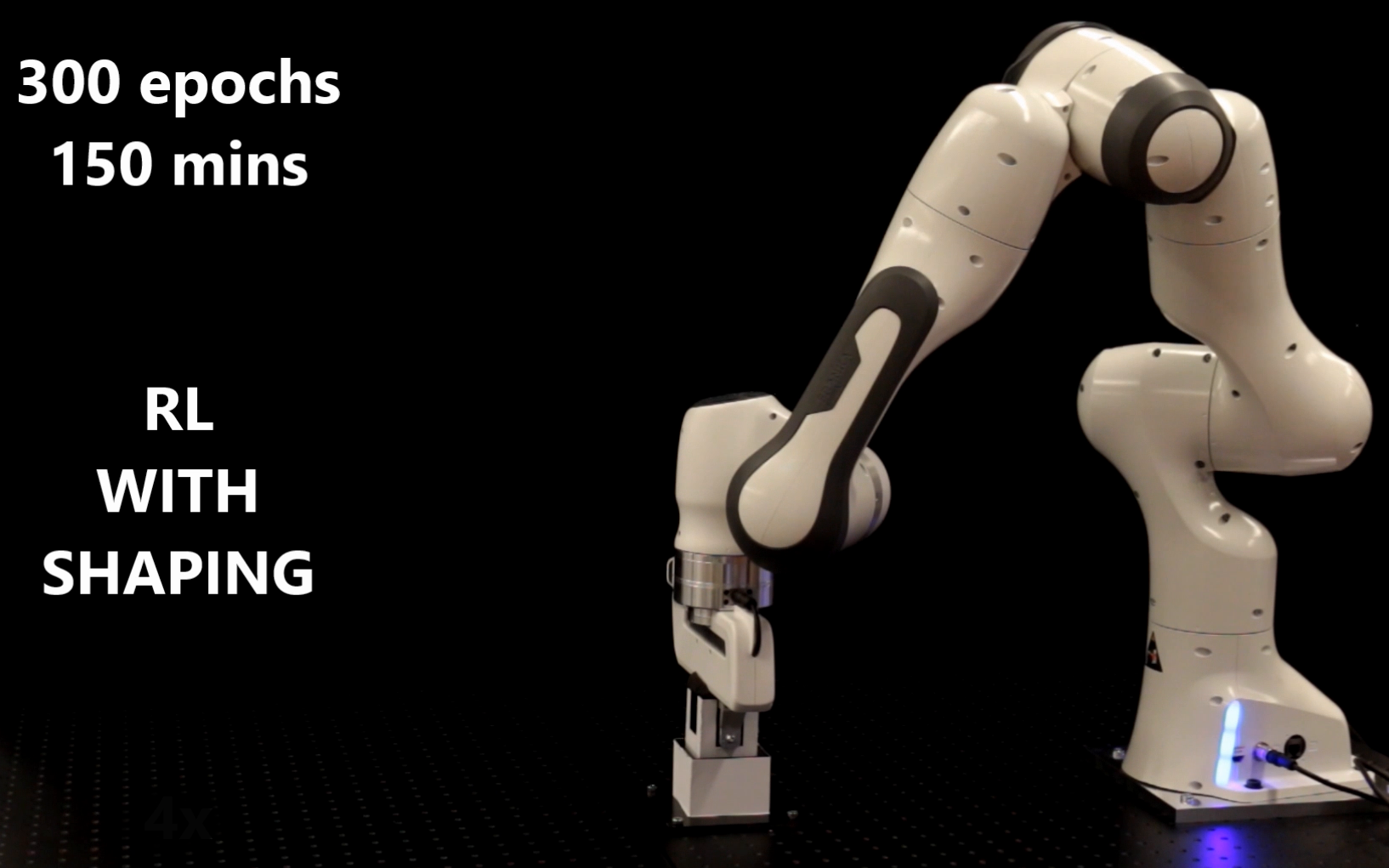} \\

\end{tabular}
\end{center}
\label{table:sample_from_video}
\caption{Snapshots of the trained peg in hole policy after 60, 120, and 300 epochs, at the end of policy execution. On the top row, the policy has been trained using RL. On the bottom row, the policy has been trained through our method, RL and reward shaping.} 
\vspace{-0.5cm}
\end{figure}

\begin{figure}[h!]
    \centering
    \includegraphics[width=0.4\textwidth]{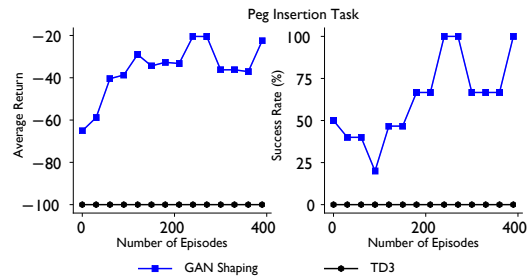}
    \vspace{-0.4cm}
    \caption{Comparison of TD3 with TD3 with shaping from a GAN potential, trained from demonstrations on a \textit{peg insertion} task. Our method finds a good policy in about 200 episodes. The performance reduction after that is due to the RL objective trying to optimize the trajectory of the peg so as to minimize time to arrive to the bottom of the hole. As it tries to optimize and straighten the trajectory, the peg starts hitting the holder more frequently, which delays the learning. To address this the reward can be modified to penalize contact with the holder.}
    \label{fig:realrobotexps2smooth}
\end{figure}

The success rates of our method and the baselines on the \textit{peg insertion} task on the real robot arm are presented in Fig.~\ref{fig:realrobotexps2smooth}, where we compare pure RL and RL with GAN shaping. The failure of TD3 to discover the goal area inside the peg holder is not surprising given the long horizon and sparse rewards involved in the task. To generate demonstration data, a near-optimal predefined trajectory was used.

Fig.~\ref{fig:realrobotexps2smooth} shows the average return from 5 episodes. Since the episode length is set to $100$, and the agent receives $-1$ when the peg is not above or in the hole, an average reward of $-100$ means the robot received no reward throughout the entire episode. We can see that with our method, RL with GAN Shaping, the robot is able to collect rewards in $20$ steps. Note that the agent does not have access to this cumulative, dense reward during training. This dense return is used here for evaluation purposes only. 



\vspace{-0.2cm}






\vspace{-0.5cm}
\section{CONCLUSION}
We addressed the problem of combining reinforcement learning with suboptimal demonstrations, using results from reward shaping and state-action potentials in order to model the demonstration data as \textit{advice}, and not as a set of \textit{constraints}, which is a popular method currently used in practice. We modeled the demonstration data as deep generative models, based on normalizing flows and Generative Adversarial Networks, and we showed that RL with generative model potentials is typically more robust than RL with behavioral cloning constraints, even in the presence of suboptimal data. We showed that our method is practical on a real robot arm, in addition to validating our method in simulation.







\bibliography{main.bib}{}
\bibliographystyle{IEEEtran}

\end{document}